\documentclass{llncs}

\usepackage{amsmath,amsfonts,amssymb,amscd}
\usepackage{graphics}
\usepackage{epsfig}
\usepackage{mathrsfs}

\usepackage{latexsym}
\usepackage{epsfig}

\usepackage{txfonts}

\newcommand{\e}{{\epsilon}}

\newcommand{\VC}{{\mbox{VC}}}
\newcommand{\abs}[1]{\lvert#1\rvert}

\def\N{{\mathbb N}}

\def\R{{\mathbb R}}
\def\modd{\,{\mathrm{mod}}\,}

\begin{document}

\begin{frontmatter}

\title{PAC learnability of a concept class under non-atomic measures: a problem by Vidyasagar}

\titlerunning{Learnability under non-atomic measures}

\author{Vladimir Pestov}
\authorrunning{V. Pestov}

\institute{Department of Mathematics and Statistics, University of Ottawa, 585 King Edward Avenue, Ottawa, Ontario, Canada K1N 6N5} 

\maketitle

\begin{abstract}
In response to a 1997 problem of M. Vidyasagar, we state a necessary and sufficient condition for distribution-free PAC learnability of a concept class $\mathscr C$ under the family of all non-atomic (diffuse) measures on the domain $\Omega$. Clearly, finiteness of the classical Vapnik--Chervonenkis dimension of $\mathscr C$ is a sufficient, but no longer necessary, condition. Besides, learnability of $\mathscr C$ under non-atomic measures does not imply the uniform Glivenko--Cantelli property with regard to non-atomic measures.
Our learnability criterion is stated in terms of a combinatorial parameter $\VC({\mathscr C}\,{\mathrm{mod}}\,\omega_1)$ which we call the VC dimension of $\mathscr C$ modulo countable sets. The new parameter is obtained by ``thickening up'' single points in the definition of VC dimension to uncountable ``clusters''. Equivalently, $\VC(\mathscr C\modd\omega_1)\leq d$ if and only if every countable subclass of $\mathscr C$ has VC dimension $\leq d$ outside a countable subset of $\Omega$. The new parameter can be also expressed as the classical VC dimension of $\mathscr C$ calculated on a suitable subset of a compactification of $\Omega$. We do not make any measurability assumptions on $\mathscr C$, assuming instead the validity of Martin's Axiom (MA).
\end{abstract}

\end{frontmatter}

\section{Introduction}
A fundamental result of statistical learning theory says that for a concept class $\mathscr C$ the three conditions are equivalent: (1) $\mathscr C$ is distribution-free PAC learnable over the family $P(\Omega)$ of all probability measures on the domain $\Omega$, (2) $\mathscr C$ is a uniform Glivenko--Cantelli class with regard to $P(\Omega)$, and (3) the Vapnik--Chervonenkis dimension of $\mathscr C$ is finite \cite{VC1971,BEHW}.
In this paper we are interested in the problem, discussed by Vidyasagar in both editions of his book \cite{vidyasagar1997,vidyasagar2003} as problem 12.8, of giving a similar combinatorial description of concept classes $\mathscr C$ which are PAC learnable under the family $P_{na}(\Omega)$ of all non-atomic probability measures on $\Omega$. (A measure $\mu$ is {\em non-atomic}, or {\em diffuse,} if every set $A$ of strictly positive measure contains a subset $B$ with $0<\mu(B)<\mu(A)$.)

The condition $\VC({\mathscr C})<\infty$, while of course sufficient for $\mathscr C$ to be learnable under $P_{na}(\Omega)$,
is not necessary. 
Let a concept class $\mathscr C$ consist of all finite and all cofinite subsets of a standard Borel space $\Omega$. Then $\VC({\mathscr C})=\infty$, and moreover $\mathscr C$ is clearly not a uniform Glivenko-Cantelli class {\em with regard to non-atomic measures.} At the same time, $\mathscr C$ is PAC learnable under non-atomic measures: any learning rule $\mathcal L$ consistent with the subclass $\{\emptyset,\Omega\}$ will learn $\mathscr C$. Notice that $\mathscr C$ is not {\em consistently} learnable under non-atomic measures: there are consistent learning rules mapping every training sample to a finite set, and they will not learn any cofinite subset of $\Omega$. 

The point of this example is that PAC learnability of a concept class $\mathscr C$ under non-atomic measures is not affected by adding to $\mathscr C$ symmetric differences $C\bigtriangleup N$ for each $C\in {\mathscr C}$ and every 
countable set $N$.

\begin{figure}[ht]
\begin{center}
\scalebox{0.225}[0.225]{\includegraphics{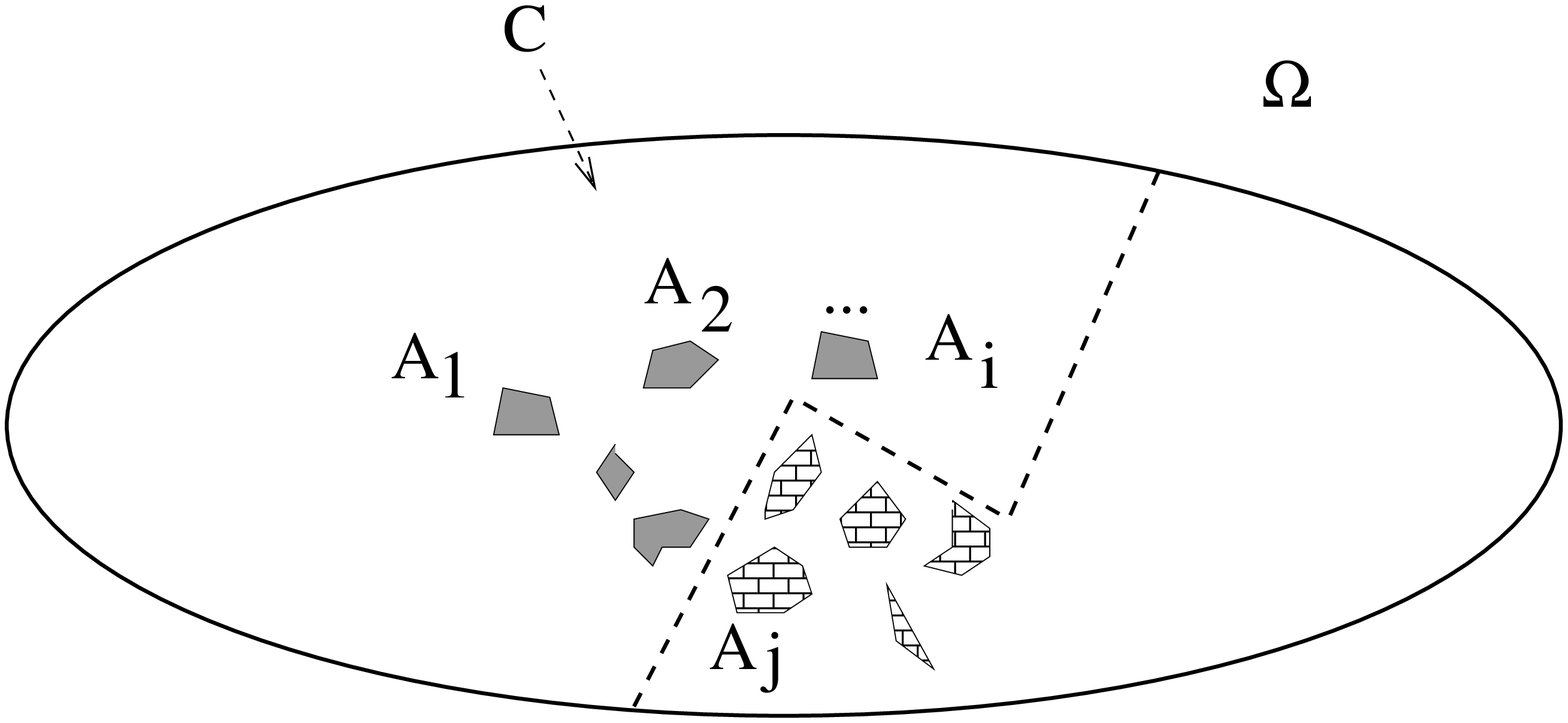}} 
\caption{A family $A_1,A_2,\ldots,A_n$ of uncountable sets shattered by $\mathscr C$.}
\label{fig:shattered-3}
\end{center}
\end{figure}

A version of VC dimension oblivious to this kind of set-theoretic ``noise'' is obtained from the classical definition by ``thickening up'' individual points and replacing them with uncountable clusters (Figure \ref{fig:shattered-3}).

Define the {\em VC dimension of a concept class $\mathscr C$ modulo countable sets} as the supremum of natural $n$ for which there exists a family of $n$ uncountable sets, $A_1,A_2,\ldots,A_n\subseteq\Omega$, shattered by $\mathscr C$ in the sense that for each $J\subseteq \{1,2,\ldots,n\}$, there is $C\in {\mathscr C}$ 
which contains all sets $A_i$, $i\in J$, and is disjoint from all sets $A_j$, $j\notin J$. Denote this parameter by $\VC({\mathscr C}\modd\omega_1)$. Clearly, for every concept class $\mathscr C$ 
\[\VC({\mathscr C}\modd\omega_1)\leq \VC({\mathscr C}).\]
In our example above, one has $\VC({\mathscr C}\modd\omega_1)=1$, even as $\VC({\mathscr C})=\infty$.

Here is our main result.

\begin{theorem}
\label{th:main}
Let $(\Omega,{\mathscr A})$ be a standard Borel space, and let ${\mathscr C}\subseteq {\mathscr A}$ be a concept class. Under the Martin's Axiom (MA), the following are equivalent.

\begin{enumerate}
\item \label{mainth:1} 
$\mathscr C$ is PAC learnable under the family of all non-atomic measures.
\item \label{mainth:3} 
$\VC({\mathscr C}\modd\omega_1)=d<\infty$. 
\item \label{mainth:4} 
Every countable subclass ${\mathscr C}^\prime\subseteq {\mathscr C}$ has finite VC dimension on the complement to some countable subset of $\Omega$ (which depends on ${\mathscr C}^\prime$).
\item\label{mainth:4a} There is $d$ such that for every countable ${\mathscr C}^\prime\subseteq {\mathscr C}$ one has $\VC({\mathscr C}^\prime)\leq d$ on the complement to some countable subset of $\Omega$ (depending on $\mathscr C^\prime$). 
\item\label{mainth:5a} Every countable subclass ${\mathscr C}^\prime\subseteq {\mathscr C}$ is a uniform Glivenko--Cantelli class with regard to the family of non-atomic measures.
\item\label{mainth:5} Same, with sample complexity $s(\e,\delta)$ which only depends on $\mathscr C$ and not on ${\mathscr C}^\prime$.
\end{enumerate}
If $\mathscr C$ is universally separable \cite{pollard}, the above are also equivalent to:
\begin{enumerate}
\item[7.] $\VC$ dimension of $\mathscr C$ is finite outside of a countable subset of $\Omega$.
\item[8.] $\mathscr C$ is a uniform Glivenko-Cantelli class with respect to the family of non-atomic probability measures.
\end{enumerate}
\end{theorem}

Martin's Axiom (MA) \cite{fremlin} is one of the most often used and best studied additional set-theoretic assumptions beyond the standard Zermelo-Frenkel set theory with the Axiom of Choice (ZFC). In particular, Martin's Axiom follows from the Continuum Hypothesis (CH), but it is also compatible with the negation of CH, and in fact it is namely the combination MA+$\neg$CH that is really interesting. 

The concept class in our initial simple example (which is even image admissible Souslin \cite{dudley}) shows that in general (7) and (8) are not equivalent to the remaining conditions. Notice that for universally separable classes, (\ref{mainth:1}), (7) and (8) are equivalent without additional set-theoretic assumptions.

The core of the theorem --- and the main technical novelty of our paper --- is the proof of the implication (\ref{mainth:4})$\Rightarrow$(\ref{mainth:1}). It is based on a special choice of a consistent learning rule $\mathcal L$ having the property that for every concept $C\in{\mathscr C}$, the image of all learning samples of the form $(\sigma,C\cap\sigma)$ under $\mathcal L$ forms a uniform Glivenko--Cantelli class. It is for establishing this property of $\mathcal L$ that we need Martin's Axiom.

Most of the remaining implications are relavely straightforward adaptations of the standard techniques of statistical learning. Nevertheless, 
(\ref{mainth:3})$\Rightarrow$(\ref{mainth:4}) requires a certain technical dexterity, and we study this implication in the setting of Boolean algebras.

We begin the paper by reviewing a general formal setting, followed by a dicussion of Boolean algebras which seem like a natural framework for the problem at hand, especially in view of possible generalizations to learning under other intermediate families of measures.

In particular, we will show that our version of the VC dimension modulo countable sets, $\VC({\mathscr C}\modd\omega_1)$, is just the usual VC dimension of the class $\mathscr C$ of concepts extended over a suitable compactification of $\Omega$ and restricted to a certain subdomain of the compactification. 

Now the part of Theorem \ref{th:main} for universally separable concept classes follows easily. Afterwards, we discuss Martin's Axiom, prove the existence of a learning rule with the above special property, and deduce 
Theorem \ref{th:main} for arbitrary concept classes. 

\section{The setting}

We need to fix a precise setting, which is mostly standard.
The {\em domain} ({\em instance space}) $\Omega=(\Omega,{\mathscr A})$ is a {\em measurable space,} that is, a set $\Omega$ equipped with a sigma-algebra of subsets $\mathscr A$. Typically, $\Omega$ is assumed to be a {\em standard Borel space,} that is, a complete separable metric space equipped with the sigma-algebra of Borel subsets. We will clarify the assumption whenever necessary. 

A {\em concept class} is a family, $\mathscr C$, of measurable subsets of $\Omega$. (Equivalently, $\mathscr C$ can be viewed as a family of measurable $\{0,1\}$-valued functions on $\Omega$.) 

In the learning model, a set $\mathcal P$ of probability measures on $\Omega$ is fixed. Usually either ${\mathcal P}=P(\Omega)$ is the set of all probability measures (distribution-free learning), or ${\mathcal P}=\{\mu\}$ is a single measure (learning under fixed distribution). In our article, the case of interest is the family ${\mathcal P}=P_{na}(\Omega)$ of all non-atomic measures.

Every probability measure $\mu$ on $\Omega$ defines a distance $d_{\mu}$ on $\mathscr A$ as follows:
\[d_{\mu}(A,B)=\mu\left(A\bigtriangleup B \right).\] 
We will not distinguish between a measure $\mu$ and its Lebesgue completion, that is, an extension of $\mu$ over the larger sigma-algebra of Lebesgue measurable subsets of $\Omega$. Consequently, we will sometimes use the term {\em measurability} meaning {\em Lebesgue measurability}. No confusion can arise here.

Often it is convenient to approximate the concepts from $\mathscr C$ with elements of the {\em hypothesis space,} $\mathscr H$, which is, technically, a subfamily of $\mathscr A$ whose closure with regard to each (pseudo)metric $d_{\mu}$, $\mu\in{\mathcal P}$, contains $\mathscr C$. However, in our article we make no distinction between $\mathscr H$ and $\mathscr C$.

A {\em learning sample} is a pair $s=(\sigma,\tau)$ of finite subsets of $\Omega$, where $\tau\subseteq\sigma$. It is convenient to assume that elements $x_1,x_2,\ldots,x_n\in\sigma$ are ordered, and thus the set of all samples $(\sigma,\tau)$ with $\abs\sigma=n$ can be identified with $\left(\Omega\times\{0,1\}\right)^n$.
A {\em learning rule} (for $\mathscr C$) is a mapping 
\[{\mathcal L}\colon \bigcup_{n=1}^\infty\Omega^n\times\{0,1\}^n\to {\mathscr C}\]
which satisfies the following measurability condition:  for every $C\in{\mathscr C}$ and $\mu\in{\mathcal L}$, the function
\begin{equation}
\label{eq:measurabilityl}
\Omega\ni\sigma\mapsto \mu\left({\mathcal L(\sigma,C\cap\sigma)}\bigtriangleup C\right) \in \R 
\end{equation}
is measurable.

A learning rule $\mathcal L$ is {\em consistent} (with $\mathscr C$) if for every $C\in {\mathscr C}$ and each $\sigma\in\Omega^n$ one has 
\[{\mathcal L}(\sigma,C\cap \sigma)\cap\sigma = C\cap\sigma.\]
A learning rule $\mathcal L$ is {\em probably approximately correct} ({\em PAC}) {\em under ${\mathcal P}$} if for every $\e>0$
\begin{equation}
\label{eq:pac}
\sup_{\mu\in {\mathcal P}}\sup_{C\in {\mathscr C}}
\mu^{\otimes n}\left\{\sigma\in\Omega^n\colon \mu\left({\mathcal L(\sigma,C\cap\sigma)}\bigtriangleup C\right)>\e \right\} \to 0\mbox{ as }n\to\infty.\end{equation}
Here $\mu^{\otimes n}$ denotes the (Lebesgue extension of the) product measure on $\Omega^n$. Now the origin of the measurability condition (\ref{eq:measurabilityl}) on the mapping $\mathcal L$ is clear: it is implicit in (\ref{eq:pac}).

Equivalently, there is a function $s(\e,\delta)$ ({\em sample complexity} of $\mathcal L$) such that for each $C\in{\mathscr C}$ and every $\mu\in {\mathcal P}$ an i.i.d. sample $\sigma$ with $\geq s(\e,\delta)$ points has the property $\mu(C\bigtriangleup {\mathcal L}(\sigma,C\cap\sigma))<\e$ with confidence $\geq 1-\delta$.

A concept class $\mathscr C$ consisting of measurable sets is {\em PAC learnable} {\em under $\mathcal P$}, if there exists a PAC learning rule for $\mathscr C$ under $\mathcal P$. A class $\mathscr C$ is {\em consistently learnable} (under $\mathcal P$) if every learning rule consistent with $\mathscr C$ is PAC under $\mathcal P$. 
If $\mathcal P = P(\Omega)$ is the set of all probability measures, then $\mathscr C$ is said to be (distribution-free) {\em PAC learnable}. At the same time, learnability under intermediate families of measures on $\Omega$ has received considerable attention, cf. Chapter 7 in \cite{vidyasagar2003}.

Notice that in this paper, we only talk of {\em potential} PAC learnability, adopting a purely information-theoretic viewpoint.

A closely related concept 
is that of a {\em uniform Glivenko--Cantelli} concept class {\em with regard to a family of measures} $\mathcal P$, that is, a concept class $\mathscr C$ such that for each $\e>0$
\begin{equation}
\label{eq:glivenko}
\sup_{\mu\in {\mathcal P}}\mu^{\otimes n}\left\{\sup_{C\in{\mathscr C}}\left\vert \mu(C)-  \mu_n(C) \right\vert\geq \e\right\}\to 0\mbox{ as }n\to\infty.
\end{equation}
(Cf. \cite{dudley}, Ch. 3; \cite{mendelson:03}.)
Here $\mu_n$ stands for the empirical (uniform) measure on $n$ points, sampled in an i.i.d. fashion from $\Omega$ according to the distribution $\mu$. One also says that $\mathscr C$ has the property of {\em uniform convergence of empirical measures} ({\em UCEM} property) {\em with regard to} $\mathcal P$ \cite{vidyasagar2003}. 

Every uniform Glivenko--Cantelli class (with regard to $\mathcal P$) is PAC learnable (under $\mathcal P$), and in the distribution-free situation, the converse is true as well.
Already in the case of learning under a single measure, it is not so: a PAC learnable class under a single distribution $\mu$ need not be uniform Glivenko-Cantelli with regard to $\mu$ (cf. Chapter 6 in \cite{vidyasagar2003}). Not every PAC learnable class under non-atomic measures is uniform Glivenko--Cantelli with regard to non-atomic measures either: the class consisting of all finite and all cofinite subsets of $\Omega$ is a counter-example.

We say, following Pollard \cite{pollard}, that a concept class $\mathscr C$ consisting of measurable sets is {\em universally separable} if it contains a countable subfamily $\mathscr C^\prime$ with the property that every $C\in {\mathscr C}$ is a pointwise limit of a suitable sequence $(C_n)_{n=1}^\infty$ of sets from $\mathscr C^\prime$: for every $x\in\Omega$ there is $N$ with the property that, for all $n\geq N$, $x\in C_n$ if $x\in C$, and $x\notin C_n$ if $x\in C$. Such a family ${\mathscr C}^\prime$ is said to be {\em universally dense} in $\mathscr C$. 

Probably the main source of uniform Glivenko--Cantelli classes is the finiteness of VC dimension. Assume that $\mathscr C$ satisfies a suitable measurability condition, for instance, $\mathscr C$ is image admissible Souslin, or else universally separable. (In particular, a countable $\mathscr C$ satisfies either condition.) If $\VC({\mathscr C})=d <\infty$, then $\mathscr C$ is uniform Glivenko--Cantelli, with a sample complexity bound that does not depend on $\mathscr C$, but only on $\e$, $\delta$, and $d$. The following is a typical (and far from being optimal) such estimate, which can be deduced, for instance, along the lines of \cite{mendelson:03}:
\begin{equation}
\label{eq:standard}
s(\e,\delta,d)\leq
\frac{128}{\e^2}\left(d\log\left(\frac{2e^2}{\e}\log\frac{2e}{\e}\right) + \log\frac 8 {\delta}\right).
\end{equation}
For our purposes, we will fix any such bound and refer to it as a {\em ``standard''} sample complexity estimate for $s(\e,\delta,d)$. 

A subset $N\subseteq\Omega$ is {\em universal null} if for every non-atomic probability measure $\mu$ on $(\Omega,{\mathscr A})$ one has $\mu(N^\prime)=0$ for some Borel set $N^\prime$ containing $N$. Universal null Borel sets are just countable sets. 

\section{VC dimension and Boolean algebras}

Recall that a {\em Boolean algebra}, $B=\langle B,\wedge,\vee,\neg,0,1 \rangle$, consists of a set, $B$, equipped with two associative and commutative binary operations, $\wedge$ (``meet'') and $\vee$ (``join''), which are distributive over each other and satisfy the absorption principles $a\vee (a\wedge b)=a$, $a\wedge (a\vee b)=a$, as well as
a unary operation $\neg$ (complement), and two elements $0$ and $1$, satisfying 
$a\vee \neg a =1$, $a\wedge \neg a=0$. 

For instance, the family $2^\Omega$ of all subsets of a set $\Omega$, with the union as join, intersection as meet, the empty set as $0$ and $\Omega$ as $1$, as well as the set-theoretic complement $\neg A = A^c$, forms a Boolean algebra. In fact, every Boolean algebra can be realized as an algebra of subsets of a suitable $\Omega$. Even better, according to the Stone representation theorem, a Boolean algebra $B$ is isomorphic to the Boolean algebra formed by all open-and-closed subsets of a suitable compact space, $S(B)$, called the {\em Stone space} of $B$, where the Boolean algebra operations are interpreted set-theoretically as above. 

The space $S(B)$ can be obtained in different ways. For instance, one can think of elements of $S(B)$ as Boolean algebra homomorphisms from $B$ to the two-element Boolean algebra $\{0,1\}$ (the algebra of subsets of a singleton). In this way, $S(B)$ is a closed topological subspace of the compact zero-dimensional space $\{0,1\}^B$ with the usual Tychonoff product topology.

The Stone space of the Boolean algebra $B=2^\Omega$ is known as the {\em Stone-\v Cech compactification of $\Omega$}, and is denoted $\beta\Omega$. 
The elements of $\beta\Omega$ are {\em ultrafilters} on $\Omega$. A collection $\xi$ of non-empty subsets of $\Omega$ is an ultrafilter if it is closed under finite intersections and if for every subset $A\subseteq\Omega$ either $A\in\xi$ or $A^c\in\xi$. To every point $x\in\Omega$ there corresponds a {\em trivial} ({\em principal}) {\em ultrafilter,} $\bar x$, consisting of all sets $A$ containing $x$. However, if $\Omega$ is infinite, the Axiom of Choice assures that there exist non-principal ultrafilters on $\Omega$. Basic open sets in the space $\beta\Omega$ are of the form $\bar A = \{\zeta\in\beta\Omega\colon A\in\zeta\}$, where $A\subseteq\Omega$. It is interesting to note that each $\bar A$ is at the same time closed, and in fact $\bar A$ is the closure of $A$ in $\beta\Omega$. Moreover, every open and closed subset of $\beta\Omega$ is of the form $\bar A$.

A one-to-one correspondence between ultrafilters on $\Omega$ and Boolean algebra homomorphisms $2^{\Omega}\to \{0,1\}$ is this: think of an ultrafilter $\xi$ on $\Omega$ as its own indicator function $\chi_\xi$ on $2^{\Omega}$, sending $A\subseteq\Omega$ to $1$ if and only if $A\in\xi$. It is not difficult to verify that $\chi_\xi$ is a Boolean algebra homomorphism, and that every homomorphism arises in this way.

The book \cite{johnstone} is a standard reference to the above topics.

Given a subset $\mathscr C$ of a Boolean algebra $B$, and a subset $X$ of the Stone space $S(B)$, one can regard $\mathscr C$ as a set of binary functions restricted to $X$, and compute the VC dimension of $\mathscr C$ over $X$. We will denote this parameter $\VC({\mathscr C}\upharpoonright X)$. 

A subset $I$ of a Boolean algebra $B$ is an {\em ideal} if, whenever $x,y\in I$ and $a\in B$, one has $x\vee y\in I$ and $a\wedge x\in I$. Define a {\em symmetric difference} on $B$ by the formula $x\bigtriangleup y =(x\vee y)\vee\neg(x\wedge y)$. 
The {\em quotient Boolean algebra} $B/I$ consists of all equivalence classes modulo the equivalence relation $x\sim y\iff x\bigtriangleup y\in I$. It can be easily verified to be a Boolean algebra on its own, with operations induced from $B$ in a unique way. 

The Stone space of $B/I$ can be identified with a compact topological subspace of $S(B)$, consisting of all homomorphisms $B\to \{0,1\}$ whose kernel contains $I$. For instance, if $B=2^{\Omega}$ and $I$ is an ideal of subsets of $\Omega$, then the Stone space of $2^{\Omega}/I$ is easily seen to consist of all ultrafilters on $\Omega$ which do not contain sets from $I$.

\begin{theorem}
\label{th:shatter}
Let $\mathscr C$ be a concept class on a domain $\Omega$, and let $I$ be an ideal of sets on $\Omega$. The following conditions are equivalent.
\begin{enumerate}
\item 
\label{i:vc}
The $VC$ dimension of the (family of closures of the) concept class $\mathscr C$ restricted to the Stone space of the quotient algebra $2^\Omega/I$ is at least $n$: $\VC({\mathscr C}\upharpoonright S(2^\Omega/I))\geq n$.
\item 
\label{i:shattered}
There exists a family $A_1,A_2,\ldots,A_n$ of measurable subsets of $\Omega$ not belonging to $I$, which is shattered by $\mathscr C$ in the sense that if $J\subseteq \{1,2,\ldots,n\}$, then there is $C\in {\mathscr C}$ which contains all sets $A_i$, $i\in J$, and is disjoint from all sets $A_i$, $i\notin J$. 
\end{enumerate}
\end{theorem}

\begin{proof}
(\ref{i:vc})$\Rightarrow$(\ref{i:shattered}). Choose ultrafilters $\xi_1,\ldots,\xi_n$ in the Stone space of the Boolean algebra $2^\Omega/I$, whose collection is shattered by $\mathscr C$. 
For every $J\subseteq \{1,2,\ldots,n\}$, select $C_J\in {\mathscr C}$ which carves the subset $\{\xi_i\colon i\in J\}$ out of $\{\xi_1,\ldots,\xi_n\}$. This means ${C_J}\in\xi_i$ if and only if $i\in J$. For all $i=1,2,\ldots,n$, set
\begin{equation}
\label{eq:bigcap}
A_i = \bigcap_{J\ni i}C_J\bigcap \bigcap_{J\not\ni i}C_J^c.
\end{equation}
Then $A_i\in\xi_i$ and hence $A_i\notin I$. Furthermore, if $i\in J$, then clearly $A_i\subseteq C_J$, and if $i\notin J$, then $A_i\cap C_J=\emptyset$.
The sets $A_i$ are measurable by their definition.
\par
(\ref{i:shattered})$\Rightarrow$(\ref{i:vc}). Let $A_1,A_2,\ldots,A_n$ be a family of subsets of $\Omega$ not belonging to the set ideal $I$ and shattered by $\mathscr C$ in sense of the lemma. For every $i$, the family of sets of the form $A_i\cap B^c$, $B\in I$ is a filter and so is contained in some free ultrafilter $\xi_i$, which is clearly disjoint from $I$ and contains $A_i$. 
If $J\subseteq \{1,2,\ldots,n\}$ and $C_J\in{\mathscr C}$ contains all sets $A_i$, $i\in J$ and is disjoint from all sets $A_i$, $i\notin J$, then the closure $\bar C_J$ of $C_J$ in the Stone space contains $\xi_i$ if and only if $i\in J$. We conclude: the collection of ultrafilters $\xi_i$, $i=1,2,\ldots,n$, which are all contained in the Stone space of $2^{\Omega}/I$, is shattered by the closed sets $\bar C_J$.
\end{proof}

It follows in particular that the VC dimension of a concept class does not change if the domain $\Omega$ is compactified.

\begin{corollary}
$\VC({\mathscr C}\upharpoonright\Omega)=\VC({\mathscr C}\upharpoonright\beta\Omega)$.
\end{corollary}

\begin{proof}
The inequality $\VC({\mathscr C}\upharpoonright\Omega)\leq\VC({\mathscr C}\upharpoonright\beta\Omega)$ is trivial. To establish the converse, assume there is a subset of $\beta\Omega$ of cardinality $n$ shattered by $\mathscr C$.
Choose sets $A_i$ as in Theorem \ref{th:shatter},(\ref{i:shattered}). Clearly, any subset of $\Omega$ meeting each $A_i$ at exactly one point is shattered by $\mathscr C$.
\end{proof}

\begin{definition}
Given a concept class $\mathscr C$ on a domain $\Omega$ and an ideal $I$ of subsets of $\Omega$, we define the VC dimension of $\mathscr C$ modulo $I$, 
\[\VC({\mathscr C}\,{\mathrm{mod}}\,I) = \VC({\mathscr C}\upharpoonright S(2^\Omega/I)).\]
That is, $\VC({\mathscr C}\,{\mathrm{mod}}\,I)\geq n$ if and only if any of the equivalent conditions of Theorem \ref{th:shatter} are met. 
\end{definition}

\begin{definition}
Let $\mathscr C$ be a concept class on a domain $\Omega$. If $I$ is the ideal of all countable subsets of $\Omega$, we denote the $\VC({\mathscr C}\,{\mathrm{mod}}\,I)$ by $\VC({\mathscr C}\modd{\omega_1})$ and call it the {\em VC dimension modulo countable sets}.
\end{definition}

\section{\label{s:necessary}Finiteness of VC dimension modulo countable sets is necessary for learnability}

\begin{lemma}
\label{l:supports}
Every uncountable Borel subset of a standard Borel space supports a non-atomic Borel probability measure.
\end{lemma}

\begin{proof}
Let $A$ be an uncountable Borel subset of a standard Borel space $\Omega$, that is, $\Omega$ is a Polish space equipped with its Borel structure. According to Souslin's theorem (see e.g. Theorem 3.2.1 in \cite{arveson}), there exists a Polish (complete separable metric) space $X$ and a continuous one-to-one mapping $f\colon X\to A$. The Polish space $X$ must be therefore uncountable, and so supports a diffuse probability measure, $\nu$. The direct image measure $f_{\ast}\nu =\nu(f^{-1}(B))$ on $\Omega$ is a Borel probability measure supported on $A$, and it is diffuse because the inverse image of every singleton is a singleton in $X$ and thus has measure zero.
\end{proof}

The following result makes no measurability assumptions on the concept class.

\begin{theorem}
\label{th:necessary}
Let $\mathscr C$ be a concept class on a domain $(\Omega,{\mathscr B})$ which is a standard Borel space. If $\mathscr C$ is PAC learnable under non-atomic measures, then the VC dimension of $\mathscr C$ modulo countable sets is finite.
\end{theorem}

\begin{proof}
This is just a minor variation of a classical result for distribution-free PAC learnability (Theorem 2.1(i) in \cite {BEHW}; we will follow the proof as presented in \cite{vidyasagar2003}, Lemma 7.2 on p. 279). 

Suppose $\VC({\mathscr C}\modd\omega_1)\geq d$. According to Theorem \ref{th:shatter}, there is a family of uncountable Borel sets $A_i$, $i=1,2,\dots, d$, shattered by $\mathscr C$ in our sense. Using Lemma \ref{l:supports}, select for every $i=1,2,\ldots,d$ a non-atomic probability measure $\mu_i$ supported on $A_i$, and let $\mu = \frac 1d\sum_{i=1}^d\mu_i$. This $\mu$ is a non-atomic Borel probability measure, giving each $A_i$ equal weight $1/d$. 

For every $d$-bit string $\sigma$ there is a concept $C_\sigma\in{\mathscr C}$ which contains all $A_{i}$ with $\sigma_i=1$ and is disjoint from $A_i$ with $\sigma_i=0$. 
If $A$ and $B$ take constant values on all the sets $A_i$, $i=1,2,\ldots,d$, then $d_{\mu}(A,B)$ is just the normalized Hamming distance between the corresponding $d$-bit strings. Now, given $A\in{\mathscr C}$ and $0\leq k\leq d$, there are 
\[\sum_{k\leq 2\e d}{d \choose k}\]
concepts $B$ with $d_{\mu}(A,B)\leq 2\e$. This allows to get the following lower bound on the number of pairwise $2\e$-separated concepts:
\[\frac {2^d}{\sum_{k\leq 2\e d}{d\choose k}}.\]
The Chernoff--Okamoto bound allows to estimate the above expression from below by $\exp[2(0.5-2\e)^2d]$. We conclude: the metric entropy of $\mathscr C$ with regard to $\mu$ is bounded below as:
\[M(2\e,{\mathscr C},\mu)\geq \exp[2(0.5-2\e)^2d].\]

The assumption $\VC({\mathscr C}\modd\omega_1)=\infty$ now implies that for every $0<\e<0.25$,
\[\sup_{P\in {\mathcal P}}M(2\e,{\mathscr C},\mu)=\infty,\]
where $\mathcal P$ denotes the family of all non-atomic measures on $\Omega$.
By Lemma 7.1 in \cite{vidyasagar2003}, p. 278, the class $\mathscr C$ is not PAC learnable under $\mathcal P$.
\end{proof}

\section{The universally separable case}

\begin{lemma}
Let $\mathscr C$ be a universally separable concept class, and let $\mathscr C^\prime$ be a universally dense countable subset of $\mathscr C$. Then
\[\VC({\mathscr C}) = \VC({\mathscr C}^\prime).\]
\label{l:shatt}
\end{lemma}

\begin{proof}
For every $C\in {\mathscr C}$ there is a sequence $(C_n)$ of elements of $\mathscr C^\prime$ with the property that for each $x\in\Omega$ there is $N$ such that if $n\geq N$ and $x\in C$, then $x\in C_n$, and if $x\notin C$, then $x\notin C_n$. Equivalently, for every finite $A\subseteq\Omega$, there is an $N$ so that whenever $n\geq N$, one has $C_n\cap A = C\cap A$. This means that if $A$ is shattered by $\mathscr C$, it is equally well shattered by $\mathscr C^\prime$. This established the inequaity $\VC({\mathscr C}) \leq \VC({\mathscr C}^\prime)$, while the converse inequality is obviously true.
\end{proof}

\begin{theorem}
\label{th:uscc}
For a universally separable concept class $\mathscr C$, the following conditions are equivalent.
\begin{enumerate}
\item\label{c:uscc1} 
$\VC({\mathscr C}\modd\omega_1)\leq d$.
\item \label{c:uscc2} 
There exists a countable subset $A\subseteq \Omega$ such that $\VC({\mathscr C}\upharpoonright(\Omega\setminus A))\leq d$.
\end{enumerate}
\end{theorem}

\begin{proof}
(\ref{c:uscc1})$\Rightarrow$(\ref{c:uscc2}): Choose a countable universally dense subfamily $\mathscr C^\prime$ of $\mathscr C$. Let $\mathscr B$ be the smallest Boolean algebra of subsets of $\Omega$ containing $\mathscr C^\prime$. Denote by $A$ the union of all elements of $\mathscr B$ that are countable sets. Clearly, $\mathscr B$ is countable, and so $A$ is a countable set. 

Let a finite set $B\subseteq\Omega\setminus A$ be shattered by $\mathscr C$. Then, by Lemma \ref{l:shatt}, it is shattered by ${\mathscr C}^\prime$. Select a family $\mathscr S$ of $2^{\abs B}$ sets in ${\mathscr C}^\prime$ shattering $B$.
For every $b\in B$ the set 
\[[b]=\bigcap_{b\in C\in {\mathscr S}} C \bigcap \bigcap_{b\notin C\in {\mathscr S}}C^c\] 
is uncountable (for it belongs to $\mathscr B$ yet is not contained in $A$), and the collection of sets $[b]$, $b\in B$ is shattered by ${\mathscr C}^\prime$. This establishes the inequality $\VC({\mathscr C}\upharpoonright (\Omega\setminus A))\leq\VC({\mathscr C}\modd{\omega_1})$. 

(\ref{c:uscc2})$\Rightarrow$(\ref{c:uscc1}): Fix an $A\subseteq\Omega$ so that $\VC({\mathscr C}\modd A^c)\leq d$. 
Suppose a collection of $n$ uncountable sets $A_i$, $i=1,2,\ldots,n$ is shattered by $\mathscr C$ in our sense. The sets $A_i\setminus A$ are non-empty; pick a representative $a_i\in A_i\setminus A$, $i=1,2,\ldots,n$. The resulting set $\{a_i\}_{i=1}^n$ is shattered by $\mathscr C$, meaning $n\leq d$.
\end{proof}

\begin{corollary}
\label{c:usl}
Let $\mathscr C$ be a universally separable concept class on a Borel domain $\Omega$. If $d=\VC({\mathscr C}\modd {\omega_1})<\infty$, then $\mathscr C$ is a universal Glivenko-Cantelli class with regard to non-atomic measures and consistently PAC learnable under non-atomic measures.
\end{corollary}

\begin{proof}
The class $\mathscr C$ has finite VC dimension in the complement to a suitable countable subset $A$ of $\Omega$, hence $\mathscr C$ is a universal Glivenko-Cantelli class (in the classical sense) in the standard Borel space $\Omega\setminus A$. But $A$ is a universal null set in $\Omega$, hence clearly $\mathscr C$ is universal Glivenko-Cantelli with regard to non-atomic measures.

The class $\mathscr C$ is distribution-free consistently PAC learnable in the domain $\Omega\setminus A$, with the standard sample complexity $s(\e,\delta,d)$. Let $\mathcal L$ be any consistent learning rule for $\mathscr C$ in $\Omega$. The restriction of $\mathcal L$ to $\Omega\setminus A$ (more exactly, to $\cup_{n=1}^\infty\left((\Omega\setminus A)^n\times\{0,1\}^n \right)$) is a consistent learning rule for $\mathscr C$ restricted to the standard Borel space $\Omega\setminus A$, and together with the fact that $A$ has measure zero with regard to any non-atomic measure, it implies that $\mathcal L$ is a PAC learning rule for $\mathscr C$ under non-atomic measures, with the same sample complexity function $s(\e,\delta,d)$.
\end{proof}

\section{Martin's Axiom and learnability}

Martin's Axiom (MA) in one of its equivalent forms says that no compact Hausdorff topological space with the countable chain condition is a union of strictly less than continuum nowhere dense subsets. Thus, it can be seen as a strengthening of the statement of the Baire Category Theorem. In particular, the Continuum Hypothesis (CH) implies MA. However, MA is compatible with the negation of CH, and this is where the most interesting applications of MA are to be found. We will be using just one particular consequence of MA.

\begin{theorem}[Martin-Solovay] 
\label{th:martin-solovay}
Let $(\Omega,\mu)$ be a standard Lebesgue non-atomic probability space. 
Under MA, 
the Lebesgue measure is $2^{\aleph_0}$-additive, that is, if $\kappa<2^{\aleph_0}$ and $A_{\alpha}$, $\alpha<\kappa$ is family of pairwise disjoint measurable sets, then $\cup_{\alpha<\kappa}A_{\alpha}$ is Lebesgue measurable and
\[\mu\left(\bigcup_{\alpha<\kappa}A_{\alpha} \right) = \sum_{\alpha<\kappa}\mu(A_{\alpha}).\]
In particular, the union of less than continuum null subsets of $\Omega$ is a null subset.
\qed
\end{theorem}

For the proof and more on MA, see \cite{kunen}, Theorem 2.21, or \cite{fremlin}, or \cite{jech}, pp. 563--565.

\begin{lemma}
\label{l:l}
Let $\mathscr C$ be an infinite concept class on a measurable space $\Omega$. Denote $\kappa=\abs{\mathscr C}$ the cardinality of $\mathscr C$. There exists a consistent learning rule $\mathcal L$ for $\mathscr C$ with the property that for every $C\in {\mathscr C}$ and each $n$, the set
\begin{eqnarray}
\label{eq:l}
\left\{{\mathcal L}(\sigma,C\cap\sigma)\colon\sigma\in \Omega^n \right\}\subseteq {\mathscr C}
\end{eqnarray}
has cardinality $<\kappa$. Under MA the rule $\mathcal L$ satisfies the measurability condition (\ref{eq:measurabilityl}).
\end{lemma}

\begin{proof}
Choose a minimal well-ordering of elements of $\mathscr C$:
\[{\mathscr C}=\{C_\alpha\colon\alpha<\kappa\},\]
and set for every $\sigma\in\Omega^n$ and $\tau\in\{0,1\}^n$ the value   ${\mathcal L}(\sigma,\tau)$ equal to $C_\beta$, where 
\[\beta = \min\{\alpha<\kappa\colon C_{\alpha}\cap\sigma=\tau\},\]
provided such a $\beta$ exists. 
Clearly, for each $\alpha<\kappa$ one has
\[{\mathcal L}(\sigma,C_{\alpha}\cap\sigma)\subseteq
\{C_\beta\colon\beta\leq\alpha\},\]
which assures (\ref{eq:l}). 
Besides, the learning rule $\mathcal L$ is consistent.

Fix $C=C_{\alpha}\in {\mathscr C}$, $\alpha<\kappa$. For every $\beta\leq\alpha$ define $D_{\beta}=\{\sigma\in\Omega^n\colon C\cap\sigma = C_{\beta}\cap\sigma\}$. The sets $D_{\beta}$ are measurable, and the function \[\Omega^n\ni\sigma\mapsto \mu({\mathcal L}(C\cap\sigma)\bigtriangleup C)\in\R\]
takes a constant value $\mu(C_{\beta}\bigtriangleup C_{\alpha})$ on each set $D_{\beta}\setminus\cup_{\gamma<\beta}D_{\gamma}$, $\beta\leq\alpha$. Such sets, as well as all their possible unions, are measurable under MA by force of Martin--Solovay's Theorem \ref{th:martin-solovay}, and their union is $\Omega^n$. This implies the condition (\ref{eq:measurabilityl}) for $\mathcal L$.
\end{proof}

We again recall that a set $A\subseteq\Omega$ is {\em absolutely null} if it is Lebesgue measurable with regard to every non-atomic Borel probability measure $\mu$ on $\Omega$ and $\mu(A)=0$. 

\begin{lemma}[Assuming MA]
\label{l:vcd}
Let $\mathscr C$ be a class of Borel subsets on a standard Borel space $\Omega$. Suppose there is a natural $d$ such
that every countable subclass ${\mathscr C}^\prime\subseteq {\mathscr C}$ has  VC dimension $\leq d$ outside of an absolutely null set (which depends on $\mathscr C$). Then every subclass of $\mathscr C$ of cardinality $<{2^{\aleph_0}}$ has the same property.
\end{lemma}

\begin{proof}
By induction on the cardinality of $\mathscr C$, which we denote $\alpha$ (notice that it never exceeds $2^{\aleph_0}$, and so the proof only makes sense under the negation of the Continuum Hypothesis). Suppose the result is true for all $\beta$, $\aleph_0\leq\beta<\alpha$. Choose a minimally well-ordered chain ${\mathscr C}_\gamma,\gamma<\alpha$ of subclasses of $\mathscr C$ whose union is $\mathscr C$. For every $\gamma$, let ${\mathcal N}_\gamma$ be a universal null subset of $\Omega$ with the property that ${\mathscr C}_{\gamma}$ has VC dimension $\leq d$ outside of ${\mathcal N}_\gamma$. Martin--Sollovay's Theorem implies that ${\mathcal N}=\cup_{\gamma<\alpha}{\mathcal N}_\gamma$ is absolutely null. Consequently, each ${\mathscr C}_{\gamma}$ has VC dimension $\leq d$ outside of $\mathcal N$, and the same applies to the union of the chain.
\end{proof}

\begin{lemma}[Assuming MA] Let $\mathscr C$ be a concept class of cardinality $\kappa=\abs{\mathscr C}<2^{\aleph_0}$ on a standard Borel space $\Omega$. If $d=\VC(\mathscr C)$ is finite, then $\mathscr C$ is a uniform Glivenko--Cantelli class, with a standard sample complexity estimate $s(\e,\delta,d)$.
\end{lemma}

\begin{proof}
A transfinite induction on $\kappa$. For $\kappa=\aleph_0$ the result is classical. Else, represent $\mathscr C$ as a union of an increasing transfinite chain of concept classes ${\mathscr C}_{\alpha}$, $\alpha<\kappa$, for each of which the statement of Lemma holds. For every $\e>0$ and $n\in\N$, the set 
\[\left\{\sigma\in\Omega^n\colon \sup_{C\in{\mathscr C}}\left\vert\mu_n(\sigma)-\mu(C)\right\vert <\e\right\} = \bigcap_{\alpha<\kappa} \left\{\sigma\in\Omega^n\colon \sup_{C\in{\mathscr C}_{\alpha}}\left\vert\mu_n(\sigma)-\mu(C)\right\vert <\e\right\}
\]
is measurable by Martin-Solovay's Theorem \ref{th:martin-solovay}. Given $\delta>0$ and $n\geq s(\e,\delta,d)$, another application of the same result leads to conclude that for every $\mu\in P(\Omega)$:
\begin{eqnarray*}
\mu^{\otimes n}\left\{\sigma\in\Omega^n\colon \sup_{C\in{\mathscr C}}\left\vert\mu_n(\sigma)-\mu(C)\right\vert <\e\right\} &=& 
\mu^{\otimes n}\left(
\bigcap_{\alpha<\kappa} \left\{\sigma\in\Omega^n\colon \sup_{C\in{\mathscr C}_{\alpha}}\left\vert\mu_n(\sigma)-\mu(C)\right\vert <\e\right\}\right) \\
&=& \inf_{\alpha<\kappa} \mu^{\otimes n}\left\{\sigma\in\Omega^n\colon \sup_{C\in{\mathscr C}_{\alpha}}\left\vert\mu_n(\sigma)-\mu(C)\right\vert <\e\right\}\\
&\geq& 1-\delta,
\end{eqnarray*}
as required.
\end{proof}

The following is an immediate consequence of two previous lemmas.

\begin{lemma}[Assuming MA]
\label{l:ugc}
Under the assumptions of Lemma \ref{l:vcd}, every subclass of $\mathscr C$ of cardinality $<{2^{\aleph_0}}$ is uniform Glivenko-Cantelli with regard to the family of non-atomic measures on $\Omega$. The sample complexity of this class is the usual sample complexity $s(\delta,\e,d)$ of concept classes of VC dimension $\leq d$.
\end{lemma}

\begin{lemma}[Assuming MA]
\label{l:paclearnable}
Let $\mathscr C$ be a concept class consisting of Borel subsets of a standard Borel space $\Omega$. Assume that for some natural $d$, every countable subclass of $\mathscr C$ has VC dimension $\leq d$ outside of some universal null subset of $\Omega$. Then the class $\mathscr C$ is PAC learnable under the family of all non-atomic measures on $\Omega$, with the usual sample complexity $s(\delta,\e)$ of distribution-free PAC learning concept classes of VC dimension $\leq d$. 
\end{lemma}

\begin{proof}
Using Lemma \ref{l:l}, choose a learning rule $\mathcal L$ for $\mathscr C$ with the property in Eq. (\ref{eq:l}). 
Since the family of all Borel subsets of $\Omega$ is well-known to have cardinality continuum, for every concept $C$ and each $n$ the cardinality of the image ${\mathscr L}_C={\mathcal L}\{C\cap\sigma\colon \sigma\in\Omega^n\}\subseteq{\mathscr C}$ is strictly less than $2^{\aleph_0}$. By Lemma \ref{l:ugc}, ${\mathscr L}_C$ is a uniform Glivenko-Cantelli class with regard to non-atomic measures on $\Omega$, satisfying the standard sample complexity bound. The proof is now concluded in a standard way. 
\end{proof}

\section{The proof of the main theorem}

(\ref{mainth:1})$\Rightarrow$(\ref{mainth:3}): this is Theorem \ref{th:necessary}.

\noindent
(\ref{mainth:3})$\Rightarrow$(\ref{mainth:4}): follows from Theorem \ref{th:uscc}.

\noindent 
(\ref{mainth:4})$\Rightarrow$(\ref{mainth:4a}): assume that for every $d$ there is a countable subclass ${\mathscr C}_d$ of $\mathscr C$ with the property that the VC dimension of ${\mathscr C}_d$ is $\geq d$ after removing any countable subset of $\Omega$. Clearly, the countable class $\cup_{d=1}^\infty {\mathscr C}_d$ will have infinite VC dimension outside of every countable subset of $\Omega$, a contradiction.

\noindent
(\ref{mainth:4a})$\Rightarrow$(\ref{mainth:5}): as a consequence of a classical result of Vapnik and Chervonenkis, every countable subclass $\mathscr C^\prime$ is universal Glivenko-Cantelli with regard to all probability measures supported outside of some countable subset of $\Omega$, and a standard bound for the sample complexity $s(\delta,\e)$ only depends on $d$, from which the statement follows.

\noindent
(\ref{mainth:5})$\Rightarrow$(\ref{mainth:5a}): trivial.

\noindent
(\ref{mainth:5a})$\Rightarrow$(\ref{mainth:4}): modelling the classical argument that the uniform Glivenko-Cantelli property implies finite VC dimension, in exactly the same spirit as in the proof of our Theorem \ref{th:necessary}, one shows that the uniform Glivenko-Cantelli property of a concept class with regard to non-atomic measures implies a finite VC dimension modulo countable sets. But for a countable (more generally, universally separable) class $\mathscr C^\prime$ this means finite VC dimension after a removal of a countable set, cf. Theorem \ref{th:uscc}.

\noindent
(\ref{mainth:4})$\Rightarrow$(\ref{mainth:1}): this is Lemma \ref{l:paclearnable}, and the only implication requiring Martin's Axiom.

The equivalence of (\ref{mainth:1}), (7) and (8) in the universally separable case follows from Theorem \ref{th:uscc} and Corollary \ref{c:usl}. 
\qed

\section{Conclusion}

We have characterized concept classes $\mathscr C$ that are distribution-free PAC learnable under the family of all non-atomic probability measures on the domain. The criterion is obtained without any measurability conditions on the concept class, but at the expense of making a set-theoretic assumption in the form of Martin's Axiom. In fact, assuming MA makes things easier, and as this axiom is very natural, perhaps it deserves its small corner within the foundations of statistical learning.

It seems that generalizing the result from concept to function classes, using a version of the fat shattering dimension modulo countable sets, will not pose particular technical difficulties, and we plan to perform this extension in a full journal version of the paper, in order to keep the conference submission short.
The Boolean algebras will however have to give way to commutative $C^\ast$-algebras \cite{arveson}.

It would be still interesting to know if the present results hold without Martin's Axiom, under the assumption that the concept class $\mathscr C$ is image admissible Souslin  (\cite{dudley}, pages 186--187). The difficulty here is selecting a measurable learning rule $\mathcal L$ with the property that the images of all learning samples $(\sigma,C\cap\sigma)$, $\sigma\in\Omega^n$, are uniform Glivenko-Cantelli. An obvious route to pursue is the recursion on the Borel rank of $\mathscr C$, but we were unable to follow it through.

Now, a concept class $\mathscr C$ will be learnable under diffuse measures provided there is a hypothesis class $\mathscr H$ which has finite VC dimension and such that every $C\in{\mathscr C}$ differs from a suitable $H\in {\mathscr H}$ by a null set. If $\mathscr C$ consists of all finite and all cofinite subsets of $\Omega$, this $\mathscr H$ is given by $\{\emptyset,\Omega\}$. One may conjecture that $\mathscr C$ is learnable under diffuse measures if and only if it admits such a ``core'' $\mathscr H$ having finite VC dimension. Is this  true?

Another natural question is: can one characterize concept classes that are uniformly Glivenko--Cantelli with regard to all non-atomic measures? Apparently, this task requires yet another version of shattering dimension, which is strictly intermediate between Talagrand's ``witness of irregularity'' \cite{talagrand96} and our VC dimension modulo countable sets. We do not have a viable candidate.

Finally, our investigation open up a possibility of linking learnability and VC dimension to Boolean algebras and their Stone spaces. This could be a glib exercise in generalization for its own sake, or maybe something deeper if one manages to invoke model theory and forcing.

\bibliographystyle{splncs}

\end{document}